\documentclass[10pt,twocolumn,letterpaper]{article}
\usepackage[accsupp]{axessibility}
\usepackage[pagenumbers]{cvpr}

\usepackage{algorithm}
\usepackage{algpseudocode}
\usepackage{bbding}
\usepackage{multirow}

\usepackage{footmisc}
\usepackage{cuted}
\definecolor{cvprblue}{rgb}{0.21,0.49,0.74}
\usepackage[pagebackref,breaklinks,colorlinks,allcolors=cvprblue]{hyperref}
\usepackage[capitalize]{cleveref}
\def\paperID{}
\def\confName{CVPR}
\def\confYear{2026}

\title{CubeComposer: Spatio-Temporal Autoregressive 4K 360° Video \\ Generation from Perspective Video}

\author{
Lingen Li$^{1}$ \quad Guangzhi Wang$^{2}$\thanks{Project lead.} \quad Xiaoyu Li$^{2}$ \quad Zhaoyang Zhang$^{2}$ \\ \quad Qi Dou$^{1}$ \quad Jinwei Gu$^{1}$ \quad Tianfan Xue$^{1}$\thanks{Corresponding author.} \quad Ying Shan$^{2}$\\
{
$^{1}$The Chinese University of Hong Kong \ 
$^{2}$ARC Lab, Tencent PCG}
}

\begin{document}
\maketitle
\begin{strip}
\vspace{-6em}
\centering
\includegraphics[width=1\linewidth]{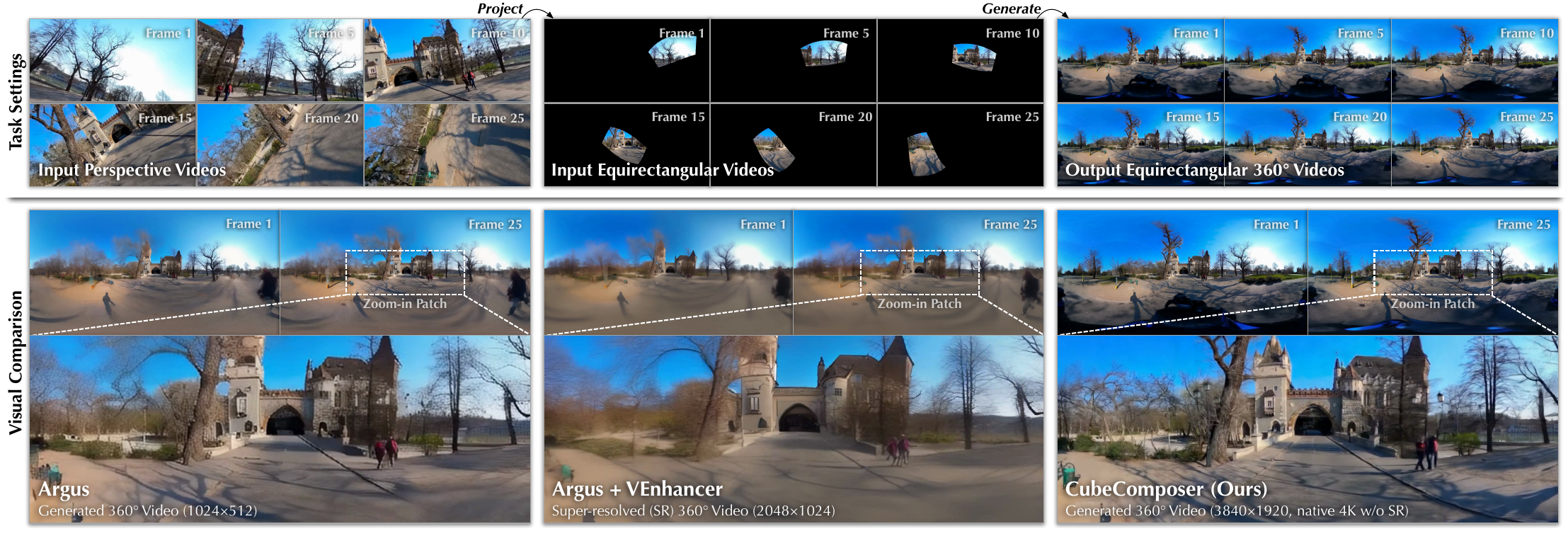}
\captionof{figure}{Existing perspective-to-360° video generation models are typically limited to a maximum resolution of 1K~\cite{tan2024imagine360,luo2025beyond,fang2025panoramic}, as they rely on the generation capability of vanilla video diffusion models with full attention. Even when augmented with advanced video super-resolution techniques like VEnhancer~\cite{he2024venhancer}, the video quality of the state-of-the-art method Argus~\cite{luo2025beyond} remains unsatisfactory. In contrast, our CubeComposer introduces a spatio-temporal autoregressive diffusion model featuring an effective context mechanism and efficient attention design, enabling for the first time the native generation (without super-resolution) of 4K 360° videos with diffusion models. \textbf{Zoom in for a better view.} \vspace{1em}}
\label{fig:teaser}
\vspace{-1em}
\end{strip}
\begin{abstract}
Generating high-quality 360° panoramic videos from perspective input is one of the crucial applications for virtual reality (VR), whereby high-resolution videos are especially important for immersive experience.
Existing methods are constrained by computational limitations of vanilla diffusion models, only supporting $\leq$ 1K resolution native generation and relying on suboptimal post super-resolution to increase resolution.
We introduce CubeComposer, a novel spatio-temporal autoregressive diffusion model that natively generates 4K-resolution 360° videos. By decomposing videos into cubemap representations with six faces, CubeComposer autoregressively synthesizes content in a well-planned spatio-temporal order, reducing memory demands while enabling high-resolution output. Specifically, to address challenges in multi-dimensional autoregression, we propose: (1) a spatio-temporal autoregressive strategy that orchestrates 360° video generation across cube faces and time windows for coherent synthesis; (2) a cube face context management mechanism, equipped with a sparse context attention design to improve efficiency; and (3) continuity-aware techniques, including cube-aware positional encoding, padding, and blending to eliminate boundary seams.
Extensive experiments on benchmark datasets demonstrate that CubeComposer outperforms state-of-the-art methods in native resolution and visual quality, supporting practical VR application scenarios. Project page: \url{https://lg-li.github.io/project/cubecomposer}.
\end{abstract}
\vspace{-1em}
    
\section{Introduction}
\label{sec:intro}
Immersive 360° panoramic videos are essential for virtual reality applications, enabling users to freely explore dynamic scenes from any viewpoint. 
However, capturing high-quality 360° panoramas requires dedicated multi-camera rigs or 360° cameras, whereas most existing videos are recorded with commodity perspective cameras.
We study perspective-to-360° video generation: given a perspective video captured by common cameras with possible rotation, the goal is to synthesize a full 360° video that faithfully completes the unobserved regions while preserving the original contexts and dynamics. 
This capability eliminates the need for specialized capture hardware and makes immersive content creation more accessible. 

\begin{figure}[tbp]
    \centering
    \includegraphics[width=1\linewidth]{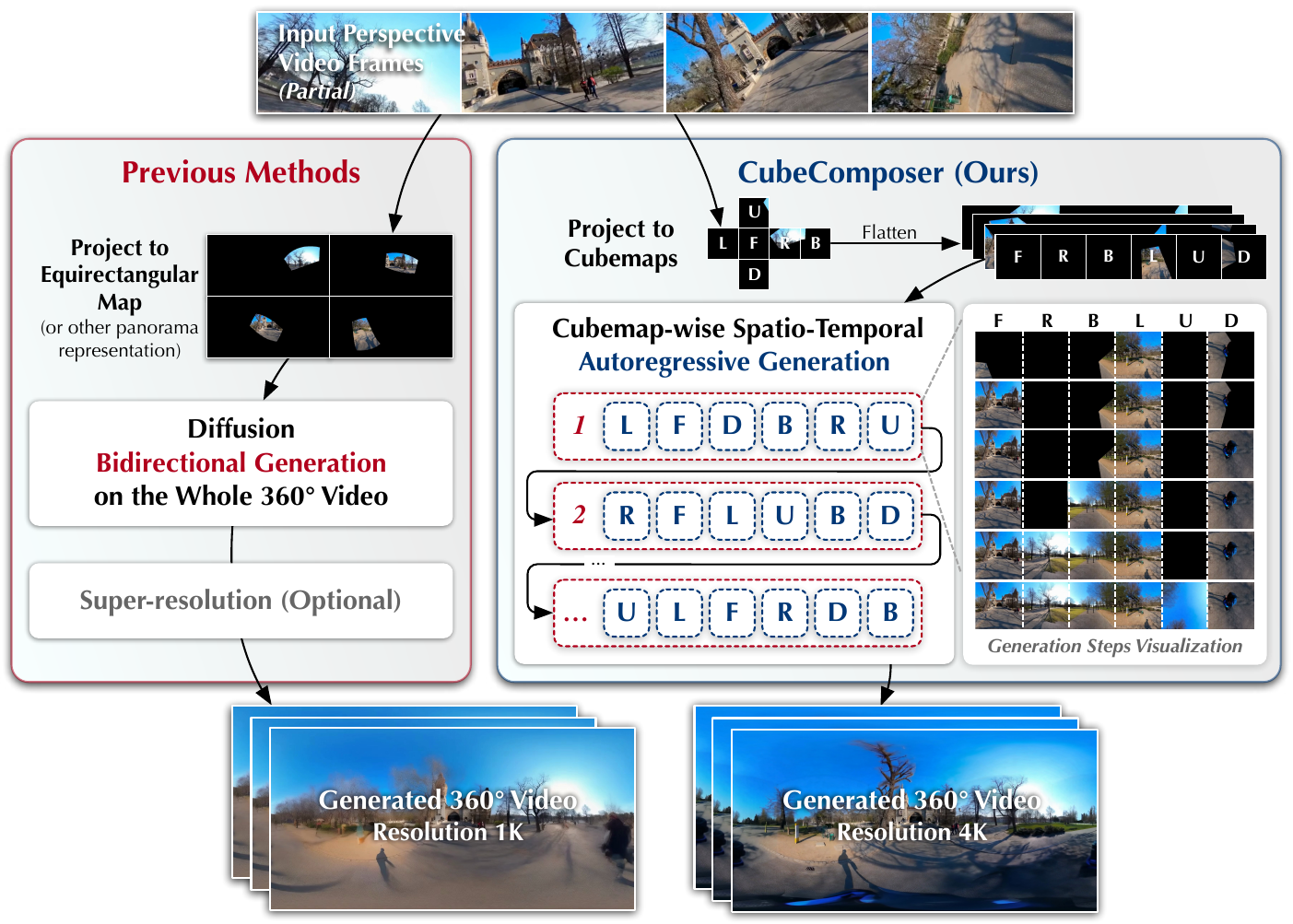}
    \caption{Comparison between the overall pipeline of previous methods and ours. CubeComposer generates the 360° video in a cubemap face-wise spatio-temporal autoregressive manner, significantly reduce the peak computational memory requirement and enable native 4K generation.}
    \label{fig:workflow-cmp}
    \vspace{-1em}
\end{figure}

Recent perspective-to-360° video methods tune foundation models~\cite{yang2024cogvideox,wan2025wan} on 360° datasets and adapt them to outpaint perspective video into equirectangular~\cite{tan2024imagine360,xie2025videopanda,luo2025beyond} or customized 360° formats~\cite{fang2025panoramic}.
Yet immersive VR experiences demand native 4K ($3840\times1920$) or even higher equirectangular resolution~\cite{lin2025one}.
Prior approaches rely on standard diffusion with full attention, incurring prohibitive computation and limiting native resolution to at most 1K ($\sim1024\times512$)~\cite{lin2025one,luo2025beyond,tan2024imagine360}, which limits generation quality and degrades user experience. To compensate for this, a super-resolution module is often added as a post-processing (\Cref{fig:workflow-cmp} left).
Such external upscaling lacks intrinsic generative reasoning and often introduces error cascades, yielding outputs that are of high resolution but deficient in details compared to native high-resolution generation, as illustrated in \Cref{fig:teaser}.

To tackle this problem, we introduce CubeComposer, a spatio-temporal autoregressive diffusion model that can generate 360° videos at 4K resolution. The key idea is to \textbf{\textit{progressively}} generate small spatio-temporal blocks one by one, rather than the entire 360° video in a single diffusion pass, which significantly reduces the peak memory. In detail, we represent a 360° video as a cubemap with six faces and synthesize it via a spatio-temporal autoregressive schedule (\Cref{fig:workflow-cmp}, right). In each step, the model generates one face over a fixed temporal window, substantially reducing peak memory usage and enabling native 4K-scale generation.

To achieve this spatial-temporal autogressive generation, our model introduces three key designs. 
First, we introduce a novel spatio-temporal generation order planning strategy tied to the perspective camera trajectory, which is causal in time and coverage-prioritized in space. Unlike temporal autoregressive models for long video extension or streaming~\cite{kim2024fifo,yin2025slow,huang2025self,lin2025autoregressive}, we also need to design a reasonable spatial generation order.
Within each time window, we compute the spatial coverage of each face and arrange their generation in descending order.
This prioritizes well-conditioned faces with more context input, which reduces early uncertainty and effectively propagates geometry, appearance, and motion cues to subsequent faces, thereby preserving cross-face coherence.

Second, to ensure the consistency of 360° video during this autoregressive generation, we design an effective and efficient context management. In each generation step, our conditional context contains: 
(1) history contents that are generated in previous temporal window; 
(2) other contents within current temporal window; and
(3) future fragments in the input perspective video related to current face for generation.
These contents provide informative context clues for generating the current face through the attention mechanism. 
To ensure efficiency, we also introduce a sparse context attention mechanism in which only the generation sequence performs full self-attention, while the context attends fully to the generation sequence locally to itself via a diagonal-banded mask, yielding linear complexity with respect to context length.

Third, autoregressively generating each cube faces may introduce discontinuities along shared boundaries, resulting in visible seams when assembled into the final 360° video. 
We mitigate this through two continuity-aware designs: 
(1) cube-aware positional encoding that incorporates the topological relationships among faces in a flattened cube layout; 
and (2) cube-aware padding and blending, where we extend each face latent with topology-aligned overlaps from adjacent faces during generation and blend the decoded overlaps in pixel space to ensure smooth transitions.

Finally, to support high-resolution training and evaluation, we curate a high-quality 360° video dataset named \textit{4K360Vid}, comprising 11,832 high-resolution 360° video clips ($\geq$ 4K). We also provide global and face-wise captions to enable optional per-face conditioning, facilitating more controlled generation in regions not covered by the input perspective video.
Extensive experiments validate the effectiveness of CubeComposer, demonstrating native 4K 360° generation with superior visual quality compared to previous methods.
Our main contributions are summarized as follows:
\begin{itemize}
    \item We present CubeComposer, the first spatio-temporal autoregressive diffusion model that natively generates 4K 360° videos from perspective inputs.
    \item We develop a 360°-video-specific autoregressive framework with a coverage-guided order tied to the input camera trajectory, enabling stable and coherent 4K 360° video generation.
    \item We propose an effective context mechanism with an efficient sparse context attention design that scales linearly with context length, improving consistency while reducing computation.
    \item We introduce continuity-aware designs, including cube-aware positional encoding and cube-aware padding and blending, to facilitate seamless boundaries.
\end{itemize}

\section{Related Work}
\label{sec:related-work}
\subsection{360° Video Generation}
Early works mainly focus on 360° image generation~\citep{akimoto2022diverse,chen2022text2light,cheng2022inout,wang2022stylelight,wu2022cross,lu2024autoregressive}. Recently, the development of video foundation models has catalyzed 360° video generation methods~\citep{lin2025one}. 
\citet{wang2024360dvd} first proposed the text and image conditioned 360° video diffusion model~\citep{wang2024360dvd}. VideoPanda~\citep{xie2025videopanda} employs multi-view attention for better consistency across different views.
In addition to text-/image-controlled 360° video generation, generation from perspective videos with varying camera rotations has also been studied in recent works.
VidPanos~\citep{ma2024vidpanos} enables panoramic outpainting with fine-tuned generative models. 
Imagine360~\citep{tan2024imagine360} supports pitch-varying perspective inputs and introduces antipodal masking for motion consistency.
Argus~\citep{luo2025beyond} enables 360° video generation from perspective inputs with varying camera rotations.
Recent works increasingly migrate video foundation models from UNet to diffusion transformers (DiT)~\citep{peebles2023scalable} backbones for better scalability and quality~\citep{zhang2025panodit,fang2025panoramic, xia2025panowan}. \citet{fang2025panoramic} trains a 360° video generation model based on a DiT backbone with a customized ViewPoint representation to improve continuity while avoiding distortion. 
However, existing methods are constrained by computational limitations of the vanilla video diffusion model, yielding native resolutions below 1K. Our work addresses this by introducing a spatio-temporal autoregressive diffusion model that natively generates 4K-resolution 360° videos.

\subsection{Video Diffusion Model}
Diffusion models have become the foundational technology in generative applications~\citep{ho2020denoising}, especially for the synthesis of images and videos~\citep{karras2022elucidating,ho2022classifier,rombach2022high,blattmann2023align,blattmann2023stable}. 
Large-scale latent video diffusion models such as Stable Video Diffusion~\citep{blattmann2023stable} demonstrate the effectiveness of learning in a compressed VAE space.
DiT-based~\citep{peebles2023scalable} architectures such as CogVideoX~\citep{yang2024cogvideox} and Wan~\citep{wan2025wan} further elevate quality through high-capacity backbones and high-quality data, yielding strong generalization across various downstream video domains.
We take advantage of rich video priors in the foundation model~\citep{wan2025wan} with our spatio-temporal autoregressive generation manner, enabling native 4K 360° video generation.

\subsection{Autoregressive Video Generation}
While most video diffusion models denoise entire clips bidirectionally in a single pass~\cite{yang2024cogvideox,wan2025wan}, recent autoregressive approaches have emerged, primarily aimed at temporal extension, streaming, and infinite generation~\citep{kim2024fifo,chen2024diffusion,guo2025long,yin2025slow,lin2025autoregressive,teng2025magi,zhang2025packing,liu2025rolling,zhuang2025flashvsr}. They typically start with a bidirectional model and convert it to an autoregressive one through training-free scheduling~\citep{kim2024fifo} or post-training with distillation~\citep{yin2025slow,lin2025autoregressive,teng2025magi}. Methods that mitigate exposure bias~\citep{huang2025self} or incorporate a better context design~\citep{zhang2025packing} further improve the quality of next-frame prediction.
Instead of extending temporally, we formulate 360° video generation as a spatio-temporal autoregressive problem over cubemap faces and time, with coverage-guided ordering and effective context mechanism.

\section{Methodology}

\begin{figure*}[htbp]
    \centering
    \includegraphics[width=1\linewidth]{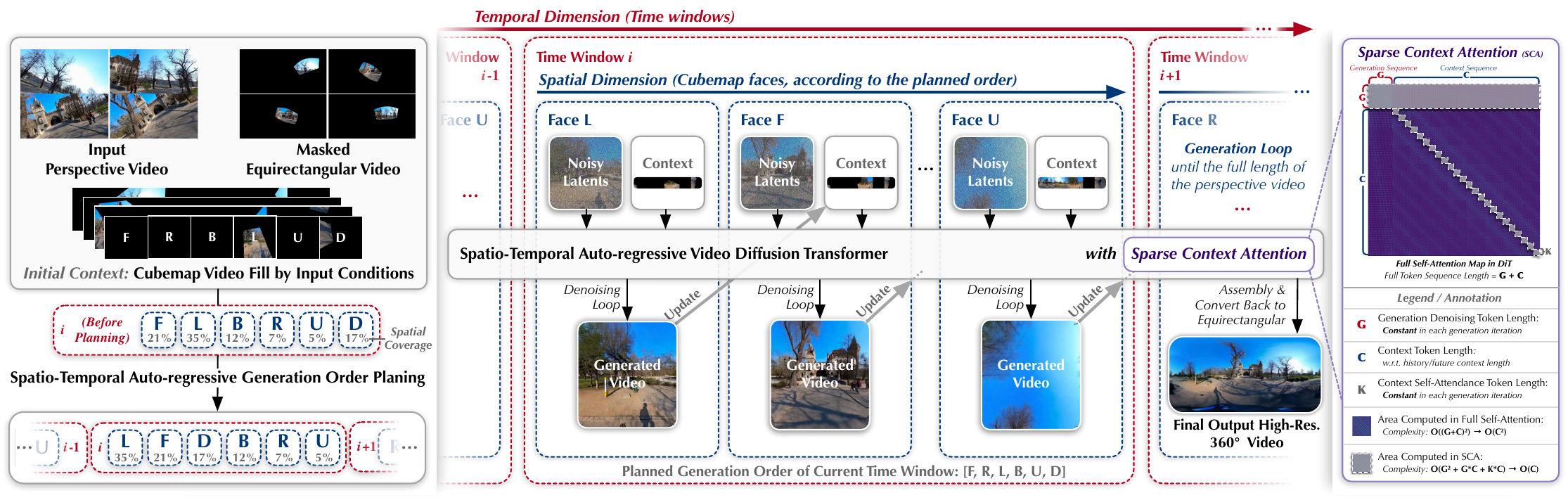}
    \caption{Pipeline overview of CubeComposer. Given a perspective video, we convert it to cubemap to obtain masked conditional inputs. The generation sequence is divided into multiple temporal windows, in which the faces are generated in a coverage-guided spatio-temporal order. At each step, CubeComposer generates a video conditioned on the context tokens with an efficient sparse context attention mechanism.
    F, R, L, B, U, D represent the front, right, left, back, up, and down faces, respectively.}
    \label{fig:method}
\end{figure*}

\begin{figure}[htbp]
    \centering
    \includegraphics[width=1\linewidth]{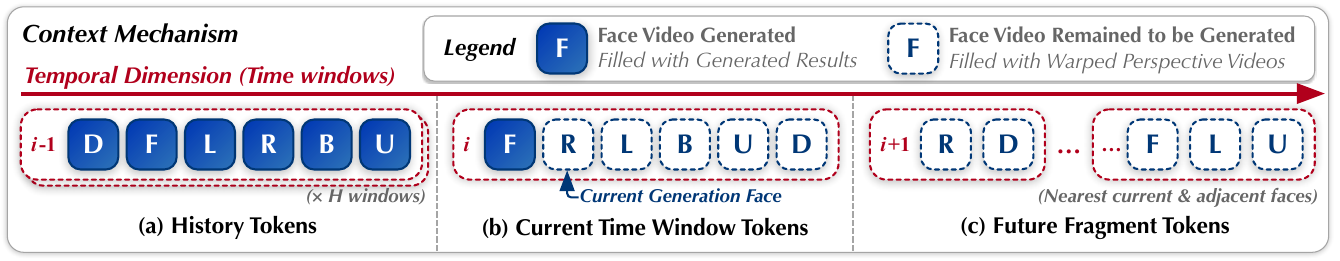}
    \caption{Context mechanism of CubeComposer, taking the generation step of face R in the $i$-th time window as example. For each generation iteration, our context mechanism composes 3 parts of tokens: (a) \textit{History Tokens}, which includes $H$ windows already generated in previous iterations; (b) \textit{Current Time Window Tokens}, which includes generated faces in the current window and perspective video conditions for ungenerated faces, always serving as a local context; and (c) \textit{Future Fragment Tokens}, where we dynamically select the temporally nearest fragment from spatially adjacent future faces (including current face) containing effective content above the spatial coverage threshold $r$.}
    \label{fig:context-mechanism}
\end{figure}

The generation of perspective-to-360° video takes a perspective video with $N$ frames as input, which is captured by a conventional camera. The output is a 360° panoramic video with the same temporal length that completing the rest angles of the scene.
To achieve this, we first estimate the camera's spherical rotations across frames and project the perspective video onto an equirectangular format, resulting in a masked equirectangular video where most regions are blank due to the limited field of view. 
We choose cubemap as the representation of panorama, since it does not apply non-uniform distortion like the equirectangular representation~\cite{huang2025dreamcube,kalischek2025cubediff}, thus better fitting the prior of existing foundation model. 
Based on this setting, the masked equirectangular video is then converted to a cubemap representation with six faces (front F, right R, back B, left L, up U, down D), serving as the conditional input.
Our model generates the complete 360° video in cubemap format, which is finally assembled into an equirectangular output, with resolution up to 4K ($3840\times1920$).

\subsection{Problem Formulation and Notation}
We denote the input perspective video by $\{I^{\mathrm{pers}}_t\}_{t=1}^{N}$, with field of view $\phi$ and per-frame camera rotation $\mathbf{R}_t \in \mathrm{SO}(3)$. Let $\mathcal{F}=\{\text{F},\text{R},\text{B},\text{L},\text{U},\text{D}\}$ be the cubemap faces and $\mathcal{N}(f)$ the set of adjacent faces of $f$. Each face has a spatial resolution of $R \times R$, and we use $P \!=\! R^2$ to denote the number of pixels per face. Projecting the perspective frames to equirectangular and converting them to cubemap yields masked conditional frames:
\begin{equation}
\label{eq:proj}
\big\{\big(X^{\mathrm{cond}}_{f,t}, M_{f,t}\big)\big\}_{f \in \mathcal{F},\, t=1}^{N}
= \Pi_{\mathrm{cube}}(I^{\mathrm{pers}}_t;\,\phi, \mathbf{R}_t),
\end{equation}
where $M_{f,t}\in\{0,1\}^{R\times R}$ is the binary mask of observed pixels and $X^{\mathrm{cond}}_{f,t}$ is the masked conditional image for face $f$ at time $t$, $\Pi_{\mathrm{cube}}$ is the projection from perspective view to the cube map representation of the 360° video.

We divide the sequence into $L$ temporal windows of equal length $T_{\mathrm{win}}$ with $L \cdot T_{\mathrm{win}} = N$, and denote the start and end of window $w$ by
\begin{equation}
\label{eq:window}
s_w = (w-1)\,T_{\mathrm{win}}, \qquad e_w = w\,T_{\mathrm{win}}.
\end{equation}
We denote $\{X^{\mathrm{cube}}_t\}_{t=1}^{N}$ as the target 360° video in cubemap format. $\{X^{\mathrm{eq}}_t\}_{t=1}^{N}$ represents the target video in equirectangular format. $\mathbf{z}_{f,t}$ means the clean latent representation of face $f$ at time $t$ and $\tilde{\mathbf{z}}_{f,t}$ denotes its generated counterpart.
Please note that equirectangular videos and cubemap videos can be converted to each others without losses. For illustration simplicity, we do not consider the temporal difference between the pixel space and the latent space of VAE in our formulation. 

\subsection{Model Overview}
 CubeComposer is a spatio-temporal autoregressive video diffusion model for perspective-to-360° video generation.
It decomposes the 360° video generation process into manageable cubemap faces and generates them autoregressively across spatial and temporal dimensions. The overall pipeline is illustrated in Figure~\ref{fig:method}.

To enable high-resolution (\eg, 4K) generation, we divide the target 360° video into $L$ temporal windows, each of  length $T_{\text{win}}$. 
Within each window, faces are generated in a planned order, conditioned on an efficient context management mechanism that incorporates historical, current, and future information with a sparse context attention design with linear complexity ($w.r.t.$ the context length).
Boundary continuity is maintained through topology-aware positional encodings, padding, and blending.
To fully take advantage of video priors, we train our model on a video foundation model Wan~\cite{wan2025wan}.

\subsection{Spatio-Temporal Autoregressive Planning}
\label{sec:method-plan}
Unlike temporal autoregression in long video generation, our task requires joint spatial and temporal autoregression to handle 360° dependencies of high resolution. 
In terms of the temporal dimension of this autoregressive process, the generation process follow a causal manner to maintain the causal consistency.
While for the spatial dimension, to maximize the fidelity with the input perspective video, we choose to generate the most certain part within each time window. This certainty can be obtained from the spatial coverage of the given perspective video in each faces.

We measure coverage as a mean over the binary mask $M_{f,t}$ for face $f$ at temporal position $t$. We use $\langle \cdot \rangle_{(i,j)}$ to denote the spatial mean over pixels. The window coverage used for order planning is the temporal mean over the window (see Eq.~\eqref{eq:window} for $[s_w,e_w)$):
\begin{equation}
\label{eq:coverage-order-cfw}
c_{f,w} \;=\; \frac{1}{T_{\mathrm{win}}}\sum_{t=s_w}^{e_w-1} \big\langle M_{f,t} \big\rangle_{(i,j)},
\end{equation}
and the within-window order sorts faces by descending coverage:
\begin{equation}
\label{eq:coverage-order}
\sigma_w \;=\; \operatorname*{argsort}_{f \in \mathcal{F}}\!\big(-c_{f,w}\big).
\end{equation}
Therefore, to determine the generation sequence, we compute $c_{f,w}$ for each face $f \in \{\text{F, R, B, L, U, D}\}$ in window $w \in \{1, \dots, L\}$ via Eq.~\eqref{eq:coverage-order}.
This prioritizes faces with more conditional information, thereby improving input fidelity and reducing error accumulation in subsequent generations. The full order is thus a sequence of $(f, s, e)$ tuples, where $s = (w-1) \cdot T_{\text{win}}$ and $e = w \cdot T_{\text{win}}$ are absolute frame indices.

\subsection{Context Mechanism with Efficient Attention}
\label{sec:method-context}
\paragraph{Context Mechanism.} To maintain coherence and consistency across face videos generated in multiple rounds, a well-designed context mechanism is essential for our spatio-temporal autoregressive diffusion model. Our task is different from long video generation where only the history context is needed, the input perspective video also provides sparse conditions across both the past and the future. For history context, we can adopt all previously generated content and set a limit to roll-out the early context. For future context, since the condition from perspective video is spatially sparse, we only need to select ``fragments'' that contain valid information instead of including all future condition that containing many empty areas.

Therefore, for each generation iteration targeting face $f$ in the $w$-th window (frame indices ranging within $[s, e)$), we let our context comprise three components, as shown in \Cref{fig:context-mechanism}: 
\begin{itemize}
    \item \textit{History tokens} $\mathbf{u}^{\mathrm{hist}}_w$: incorporate up to $H$ previous windows from generated content, stored in a context pool. 
    \item \textit{Current window tokens} $\mathbf{u}^{\mathrm{curr}}_{w,f}$: include generated faces from prior iterations in the same window and perspective conditions for ungenerated faces, including the current one.
    \item \textit{Future fragment tokens} $\mathbf{u}^{\mathrm{fut}}_{w,f}$: address temporal foresight by dynamically selecting non-consecutive segments from the perspective condition where coverage exceeds threshold $r$, focusing on the nearest available fragments from the current and adjacent faces.
\end{itemize}

We formalize the context as a concatenation of the three components along the token dimension: 
\begin{equation}
\label{eq:context}
\mathbf{u}_{w,f} = \big[ \mathbf{u}^{\mathrm{hist}}_w;\ \mathbf{u}^{\mathrm{curr}}_{w,f};\ \mathbf{u}^{\mathrm{fut}}_{w,f}\big].
\end{equation}
For future fragments, we compute a short-horizon coverage over a length $T_{\mathrm{frag}}$ starting at $\tau$:
\begin{equation}
\label{eq:short-horizon-coverage}
\bar{c}_{g}(\tau;\,T_{\mathrm{frag}}) \;=\; 
\frac{1}{T_{\mathrm{frag}}}\sum_{t=\tau}^{\tau+T_{\mathrm{frag}}-1} m_{g,t},
\end{equation}
where $g \in \mathcal{N}(f)\,\cup\,\{f\}$ indicates the face sets that containing the current face $f$ and its spatially-adjacent faces derived from $\mathcal{N}(f)$.
We select the nearest future start time whose frame's spatial short-horizon coverage exceeds the threshold $r$:
\begin{equation}
\label{eq:future-start}
\tau_g^* \;=\; \operatorname*{arg\,min}_{\tau \ge e_w} \;\big\{\, \tau \,\big|\, \bar{c}_{g}(\tau;\,T_{\mathrm{frag}}) \ge r \,\big\}.
\end{equation}
The future fragment tokens for $(w,f)$ then pack the conditional frames on adjacent faces (and the current face) over the selected short horizon:
\begin{equation}
\label{eq:future-tokens}
\mathbf{u}^{\mathrm{fut}}_{w,f} \;=\;
\Big[X^{\mathrm{cond}}_{g,\;\tau_g^*:\tau_g^*+T_{\mathrm{frag}}-1} \Big]_{g \in \mathcal{N}(f)\,\cup\,\{f\}}.
\end{equation}
Inside the DiT, the context $\mathbf{u}_{w,f}$ are encoded to latents by the VAE, tokenized with the patch embedding, and concatenated as extra tokens to the generation sequence in an in-context manner.

\paragraph{Sparse Context Attention.} 
Since the context mechanism significantly extend the token sequence during generation and the extra token length leads to a quadratic computational complexity inside the full attention of DiT models, an efficient attention design is required to handle the context part, so as to reduce the computational cost and enable higher resolution generation.

Therefore, we employ a sparse context attention design shown in the right-most part of \Cref{fig:method}. Specifically, in the self attention of CubeComposer, the generation sequence (length $G$, constant in each step) performs full self-attention, while the context sequence (length $C$) attends fully to the generation part but sparsely to itself via a diagonal-banded local mask of bandwidth $K$. This constrains context self-attention to $O(C \cdot K)$ operations.
With this design, we significantly reduce the attention computational complexity in our model from square \textit{w.r.t.} the context length $C$ to \textbf{linear}.
In our implementation, the diagonal bandwith $K$ is set to the spatial token length of a single cube face.

\begin{figure}[tbp]
    \centering
    \includegraphics[width=1\linewidth]{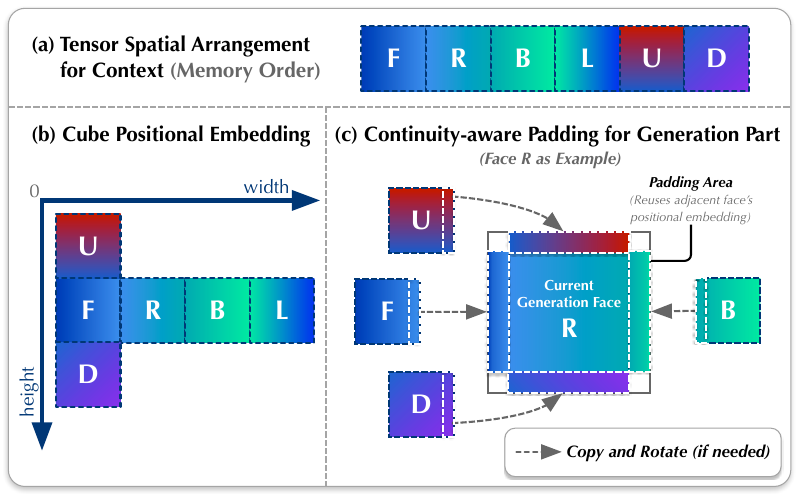}
    \caption{Continuity-aware designs in CubeComposer, which are used to tackle the discontinuity caused by the spatially separated generation in our spatio-temporal autoregressive manner.}
    \label{fig:cont-aware-designs}
    \vspace{-1em}
\end{figure}

\subsection{Continuity-aware Designs}
Generating faces autoregressively with cubemap representation and assembling them into a 360° video risks seams at cube face boundaries. To alleviate discontinuity across face boundaries, we propose a series of continuity-aware model designs, including cube-aware positional encodings and cube-aware padding blending.

Firstly, instead of applying positional encodings to latent tensors as the original spatial arrangement shown in \Cref{fig:cont-aware-designs} (a), we remap its positional encodings according to a flattened cubemap topology: U face's top index starts at 0, F face's top index starts at cube size $R$, D faces's top index starts at twice cube size $2R$, with horizontal placement for others.
Then, during generation, we pad the current face's latents and positions with strips from adjacent faces, applying rotations and flips per adjacency, as shown in \Cref{fig:cont-aware-designs} (c). After generation of current face, overlapping padding regions are blended with adjacent faces when updating the context via weighted averaging to smooth transitions.

Such a design enables effective utilization of contextual information in adjacent faces, thereby improving cross-face continuity and consistency (as validated in~\Cref{fig:ablation-continuity}).

\subsection{Training and Inference}
During training, we simulate autoregression on ground-truth 360° videos by sampling a window $w$ and face $f$, constructing the context $\mathbf{u}_{w,f}$ per \S\ref{sec:method-context}, and conditioning on the global prompt $y$ or face-wise prompt $y_f$. We train a diffusion transformer $\mathbf{v}_\theta$ to predict velocity under these conditions using the flow-matching objective~\cite{esser2024scaling}:
\begin{equation}
\label{eq:flow-matching}
\mathcal{L} = \mathbb{E}_{t \sim \mathcal{U}[0,1],\, \mathbf{z}_0 \sim p(\mathbf{z}_0)}
\left[ \left\| \mathbf{v}_\theta\big(\mathbf{z}_t, t;\, \mathbf{u}_{w,f}, y\big) - \mathbf{v}_t \right\|^2 \right],
\end{equation}
where $\mathbf{v}_\theta$ is the video DiT conditioned on the context and prompts, and $\mathbf{v}_t = \mathbf{z}_0 - \mathbf{z}_t$ denotes the velocity at time $t \in [0,1]$, $\mathbf{z}_0 $ denotes the clean latent of the target 360° video, and $\mathbf{z}_t$ denotes the noisy latent at time $t \in [0,1]$.

During inference, we perform diffusion generation on each faces while updating $\mathbf{u}_{w,f}$ according to our context mechanism, producing native 4K 360° videos without super-resolution. Please refer to the supplementary for a detailed description of the inference process.

Beyond a single global prompt $y$ used across all iterations, we want our model to support optional face-wise prompts $\{y_f\}_{f \in \mathcal{F}}$ to guide specific faces when the perspective input does not cover that area. To achieve this, we annotate data with both global and face-wise captions and randomly apply the latter during training. For inference, users may provide either a single global prompt or optional per-face prompts to control the generation.

\section{Experiments}
In this section, we first introduce the datasets we use and the training details. Then we conduct several experiments to compare the proposed CubeComposer with previous state-of-the-art perspective-to-360° video generation models to evaluate its performance. Furthermore, we conduct comprehensive ablation and analysis on model designs of CubeComposer to future verify the effectiveness of each mechanism and component design.

\begin{table*}[htbp]
\resizebox{0.98\linewidth}{!}{

\begin{tabular}{c|c|ccccccc|ccccccc}
\toprule
\multirow{2}{*}{\textbf{Model}}        & \multirow{2}{*}{\textbf{Res.}} & \multicolumn{7}{c|}{\textbf{4K360Vid Dataset}}                                       & \multicolumn{7}{c}{\textbf{ODV360 Dataset}}                                           \\ \cmidrule{3-16} 
                              &                       & LPIPS↓  & CLIP↑   & FID↓      & FVD↓     & A. Q.↑ & I. Q.↑ & O. C.↑ & LPIPS↓  & CLIP↑   & FID↓      & FVD↓     & A. Q.↑ & I. Q.↑ & O. C.↑ \\ \midrule
ViewPoint                     & 1K                  & 0.5663 & 0.8532 & 196.5319 & 3.8517  & 0.3508  & 0.3456  & 0.1699    & 0.6486 & 0.8713 & 164.4097 & 5.3720  & 0.3540  & 0.3572  & 0.1538    \\
ViewPoint+VEnhanced           & 2K                    & 0.5761 & 0.8536 & 201.8165 & 3.8522  & 0.3733  & 0.3966  & 0.1718    & 0.6339 & 0.8495 & 174.1680 & 5.7342  & 0.3625  & 0.3804  & 0.1525    \\ \midrule
Argus                         & 1K                    & 0.4074 & 0.8858 & 141.1540 & 4.0755  & 0.3715  & 0.4266  & 0.1709    & 0.4336 & 0.8794 & 140.9175 & 12.7548 & 0.3764  & 0.3988  & 0.1548    \\
Argus+VEnhanced               & 2K                    & 0.4689 & 0.8576 & 168.9571 & 6.1337  & 0.3596  & 0.4286  & 0.1639    & 0.4962 & 0.8330 & 180.6507 & 14.1573 & 0.3623  & 0.3671  & 0.1500    \\ \midrule
Imagine360                    & 1K                    & 0.7367 & 0.7930 & 254.2880 & 5.0955  & 0.3261  & 0.4576  & 0.1670    & 0.7021 & 0.8090 & 192.9936 & 9.2924  & 0.3632  & 0.4846  & \textbf{0.1660}    \\
Imagine360+VEnhanced          & 2K                    & 0.7285 & 0.7775 & 270.7605 & 10.2088 & 0.3565  & 0.4270  & 0.1505    & 0.6827 & 0.7915 & 218.0442 & 7.3203  & 0.3647  & 0.4052  & 0.1509    \\ \midrule
\multirow{2}{*}{CubeComposer} & 2K                    & \textbf{0.3696} & \textbf{0.9234} & \textbf{119.0998} & 3.9035  & 0.3984  & 0.5214  & \textbf{0.1773}    & 0.4249 & 0.8911 & 125.5510 & 4.2592  & 0.4067  & 0.5144  & 0.1616    \\
                              & 4K                    & 0.3831 & 0.9111 & 130.9209 & \textbf{2.2205}  & \textbf{0.4051}  & \textbf{0.5618}  & 0.1769    & \textbf{0.4170} & \textbf{0.9061} & \textbf{123.5605} & \textbf{3.5054}  & \textbf{0.4168}  & \textbf{0.5543}  & 0.1639    \\ \bottomrule
\end{tabular}
}
\vspace{-0.5em}
\caption{
Quantitative comparison on LPIPS, CLIP image similarity, FID, FVD, VBench~\cite{huang2024vbench} aesthetic quality (A. Q.), imaging quality (I. Q.), and overall consistency (O. C.) between CubeComposer and previous perspective-to-360° video generation methods, including ViewPoint~\cite{fang2025panoramic}, Argus~\cite{luo2025beyond}, and Imagine360~\cite{tan2024imagine360}. Pervious methods runs natively on 1K ($1024\times512$) resolution at most, add can be upscaled to 2K ($2048\times1024$) resolution with external generative superresolution model VEnhancer~\cite{he2024venhancer}.
In contrast, our CubeComposer runs natively on 2K and 4K ($3840\times1920$) resolutions. The best value in each column is noted in \textbf{bold}. For fair comparison, all metrics are calculated with ground truth with corresponding resolution, therefore models with higher resolution will be harder to achieve better metric value.
}

\label{tab:exp-main}
\end{table*}

\subsection{Experimental Settings}
\paragraph{Dataset.} 
We curate a dataset named \textit{4K360Vid}, which comprises over 11,832 high-quality 4K 360° videos. This dataset is constructed based on the public 360° video dataset used in Argus~\cite{luo2025beyond}. Each video clip is annotated with global and frame-wise captions generated by the Qwen3-VL 235B A22 Instruct model~\cite{Qwen-VL, Qwen2-VL, Qwen2.5-VL}. Additionally, we employ this vision-language model to filter out low-quality or anomalous content, ensuring high data quality. Furthermore, we use the high-resolution subset of the ODV360~\cite{cao2023ntire} dataset for training and evaluation. For full details on the datasets used, please refer to the supplementary material.

\vspace{-1em}
\paragraph{Training.}
We train CubeComposer on 4K360Vid and ODV360~\cite{cao2023ntire} with random perspective synthesis and context selection, from the foundation model Wan 2.2 5B~\cite{wan2025wan}. For each scene, we sample smooth camera trajectories (3–5 anchor points; FoV 60–120°) and follow our planned generation order and context strategy (see \Cref{sec:method-plan,sec:method-context}). Please refer to the supplementary for full training settings.

\begin{figure*}[tbp]
    \centering
    \includegraphics[width=1\linewidth]{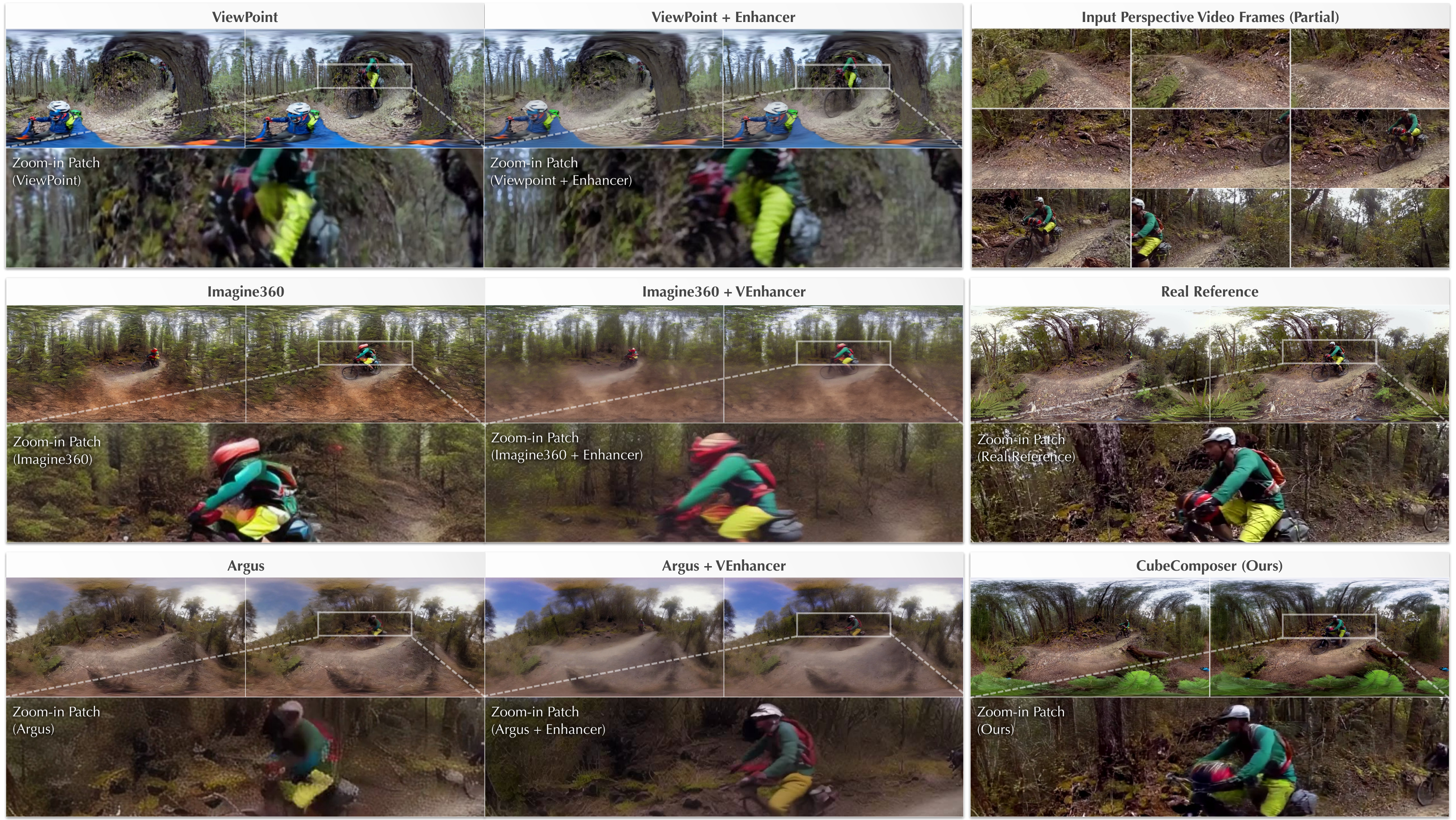}
    \vspace{-1.5em}
    \caption{Qualitative comparison of 360° video generation between ViewPoint~\cite{fang2025panoramic}, Imaging360~\cite{tan2024imagine360}, Argus~\cite{luo2025beyond}, and our CubeComposer. Our method natively generates 360° video in 4K resolution, surpassing previous methods that generates 1K resolution at most (2K with VEnhancer~\cite{he2024venhancer}) in visual quality and detail richness. \textbf{Zoom-in for a better view.} More visual results are provided in the supplementary.}
    \label{fig:exp-comp}
    \vspace{-1em}
\end{figure*}

\subsection{Comparison}

\paragraph{Baselines.} We compare our CubeComposer with the state-of-the-art 360° video generation methods Argus~\cite{luo2025beyond}, Imagine360~\cite{tan2024imagine360}, and ViewPoint~\cite{fang2025panoramic}. Since these perspective-to-video models are trained with different variety of the camera rotations, we apply perspective video input with strategies within the model's capability to ensure fairness.
For methods that support free-form camera rotation (Argus and ours), the trajectories of perspective videos are sampled from two random points (trajectories are the same for two models). 
For methods that support only limited or fixed camera rotations (Imagine360 and ViewPoint), we keep the input perspective video fixed at the front face. The horizontal and vertical input field-of-view are 90° and 45° for all methods except ViewPoint~\cite{fang2025panoramic}. Since ViewPoint only supports squared perspective input, we set its FoV to 90° for both height and width.
All previous methods are evaluated at their pre-trained maximum resolution, and we adopt the VEnhancer~\cite{he2024venhancer} as an external generative post-processor to super-resolve (2x) generated 360° videos of previous methods.
We evaluate all the models on the test set of ODV360 dataset and a randomly selected set with 20 4K 360° scenes from the 4K360Vid dataset that are unseen during training.

\vspace{-1em}
\paragraph{Metrics.} We adopt three groups of visual metrics to evaluate the performance of models in our experiments: 1) reference-based metrics LPIPS~\citep{zhang2018unreasonable} and CLIP~\citep{radford2021learning} image similarity; 2) distributional metrics including FID~\cite{heusel2017gans} for image-level calculation and FVD~\cite{unterthiner2018towards} for video-level calculation; 3) non-reference video metrics of VBench~\cite{huang2024vbench} including aesthetic quality (A. Q.), imaging quality (I. Q.), and overall consistency (O. C.). VBench metrics are calculated and averaged in six perspective projections of the generated 360° video. Since the supported resolution varies across models, we calculate the metrics under the corresponding target temporal and spatial resolution of each model (with ground truth resized to fit the target size) to ensure fairness.

\vspace{-1em}
\paragraph{Results.}
The quantitative results are shown in~\Cref{tab:exp-main} and the qualitative results are shown in~\Cref{fig:exp-comp}. The 360° videos generated by previous methods look unnatural and lack details due to the limited resolution. With the post-processing of the VEnhancer~\cite{he2024venhancer}, the unnaturalness becomes even more severe. In contrast, our method runs natively on a 4K resolution, significantly outperforming previous methods in overall visual quality and details. More video results are provided in the supplementary.

\begin{figure}
    \centering
    \includegraphics[width=1\linewidth]{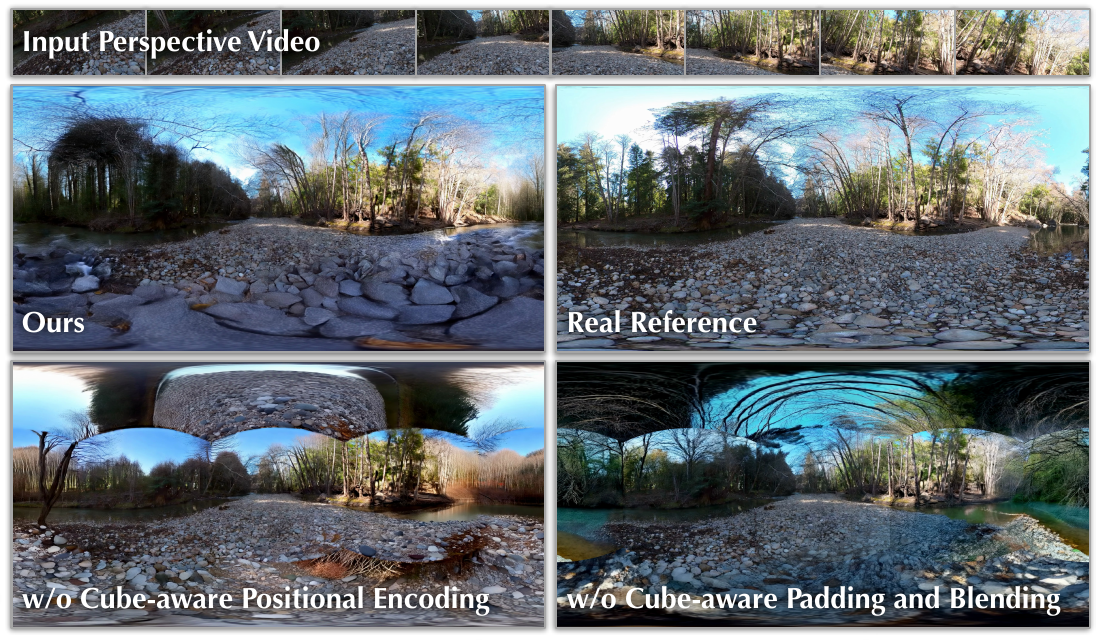}
    \vspace{-1.5em}
    \caption{Qualitative ablation of our continuity-aware designs. We compare the proposed full model (\textit{Ours}) against variants without cube-aware positional encodings (vanilla RoPE) and without padding and blending. The full model markedly reduces cross-face discontinuities in our autoregressive settings.}
    \label{fig:ablation-continuity}
    \vspace{-1em}
\end{figure}

\subsection{Ablation and Analysis}
In this section, we provide key ablation results for our model designs. Due to the space limit, more analysis, discussion, and user studies are provided in the supplementary.

\vspace{-1em}
\paragraph{Analysis on Context Designs.}
The context mechanism described in~\Cref{sec:method-context} is designed to promote consistency across spatio-temporal generation rounds. We evaluate three configurations to generate 27-frame videos on the test set of ODV360~\cite{cao2023ntire}: (1) the proposed mechanism (\textit{Ours}), (2) a full-context variant that includes all temporal tokens at each step (\textit{Full tokens}), and (3) a causal variant that excludes future tokens (\textit{w/o future tokens}). As reported in~\Cref{tab:ablation-context}, excluding future tokens substantially degrades performance. The proposed mechanism attains LPIPS, CLIP, and FID comparable to the full-context variant and achieves slightly better FVD, while incurring fewer TFLOPs. Since FVD is relatively sensitive to temporal dynamics, the lower FVD observed with our mechanism indicates stronger temporal coherence attributable to the effective context design.

\begin{table}[tbp]
    \centering
    \resizebox{\linewidth}{!}{
    \begin{tabular}{c|c|cccc}
    \toprule
    Context Type & TFLOPS & FVD↓ & FID↓ & LPIPS↓ & CLIP↑   \\ \midrule
    Ours & 350.64 & 4.2592 &  125.5510 & 0.4249 & 0.8911  \\
    w/o future tokens & 224.89  & 6.0369 & 128.3274 & 0.4517 & 0.8878  \\
    Full tokens  & 376.03 & 5.2265 & 116.6476 & 0.4162 & 0.8961   \\
       \bottomrule
    \end{tabular}
    }
    \vspace{-0.5em}
    \caption{Analysis on different choices for the context mechanism. When using the causal context manner (\textit{w/o future tokens}), we can observe a significant performance drop, indicating the importance of future tokens.
    In addition, our context mechanism is comparable to the full token model where all tokens are included as context in each step. }
    \label{tab:ablation-context}
\end{table}

\vspace{-1em}
\paragraph{Ablation on Continuity-Aware Designs.}
We assess the continuity-aware components by training three variants on the training set of ODV360 dataset for 50 epochs under identical settings from the foundation model: (1) the proposed model, (2) a model without cube-aware positional encodings (vanilla RoPE applied to the original tensor layout), and (3) a model without cube-aware padding and blending. As shown in~\Cref{tab:ablation-continuity,fig:ablation-continuity}, both components mitigate cross-face seams and temporal discontinuities in multi-step autoregressive generation. Removing either component leads to boundary artifacts and inconsistent degradations across LPIPS, CLIP, FID, and FVD. Enabling both yields the best overall performance.

\begin{table}[tbp]
    \centering
    \resizebox{\linewidth}{!}{
    \begin{tabular}{cc|cccc}
    \toprule
       Pos. Enc. & Pad. Blend. & FVD↓  & FID↓ & LPIPS↓ & CLIP↑  \\ \midrule
       \CheckmarkBold & \XSolidBrush & 4.3683  & 190.3326  & 0.5600 & 0.8409  \\ 
       \XSolidBrush   & \CheckmarkBold & 4.4650 & 201.4123  & 0.5504 & 0.8547 \\
       \CheckmarkBold & \CheckmarkBold & \textbf{4.1961} & \textbf{157.1220}  & \textbf{0.5142} & \textbf{0.8590}  \\
       \bottomrule
    \end{tabular}
    }
    \vspace{-0.5em}
    \caption{Quantitative results of ablation on the continuity-aware designs. The best result in each column is noted in bold.}
    \label{tab:ablation-continuity}
    \vspace{-1em}
\end{table}

\section{Conclusion}
We presented CubeComposer, a spatio-temporal autoregressive diffusion model that natively generates 4K 360° videos from perspective inputs without super-resolution. By representing panoramas as cubemaps and planning generation across faces and temporal windows, our approach substantially reduces peak memory while preserving global coherence. Our design combines a spatio-temporal AR generation strategy, an efficient context with sparse attention, and continuity-aware techniques to suppress seams. Experiments show consistent gains over state-of-the-art methods in native resolution, spatial seamlessness, and temporal consistency, with ablations confirming the contribution of each component. 
Looking ahead, prioritizing efficiency by reducing diffusion steps and moving towards streaming 360° generation to amortize computation and lower latency would be an interesting and promising direction.

{
    \small
    \bibliographystyle{ieeenat_fullname}
    \bibliography{main}

@String(TOG= {ACM Trans. Graph.})

@String(AAAI = {AAAI})

@String(TOG   = {ACM TOG})

@article{wan2025wan,
  title={Wan: Open and advanced large-scale video generative models},
  author={Wan, Team and Wang, Ang and Ai, Baole and Wen, Bin and Mao, Chaojie and Xie, Chen-Wei and Chen, Di and Yu, Feiwu and Zhao, Haiming and Yang, Jianxiao and others},
  journal={arXiv preprint arXiv:2503.20314},
  year={2025}
}

@article{yang2024cogvideox,
  title={Cogvideox: Text-to-video diffusion models with an expert transformer},
  author={Yang, Zhuoyi and Teng, Jiayan and Zheng, Wendi and Ding, Ming and Huang, Shiyu and Xu, Jiazheng and Yang, Yuanming and Hong, Wenyi and Zhang, Xiaohan and Feng, Guanyu and others},
  journal={arXiv preprint arXiv:2408.06072},
  year={2024}
}

@inproceedings{wang2024360dvd,
  title={360dvd: Controllable panorama video generation with 360-degree video diffusion model},
  author={Wang, Qian and Li, Weiqi and Mou, Chong and Cheng, Xinhua and Zhang, Jian},
  booktitle={Proceedings of the IEEE/CVF Conference on Computer Vision and Pattern Recognition},
  pages={6913--6923},
  year={2024}
}

@article{xie2025videopanda,
  title={VideoPanda: Video panoramic diffusion with multi-view attention},
  author={Xie, Kevin and Sabour, Amirmojtaba and Huang, Jiahui and Paschalidou, Despoina and Klar, Greg and Iqbal, Umar and Fidler, Sanja and Zeng, Xiaohui},
  journal={arXiv preprint arXiv:2504.11389},
  year={2025}
}

@article{fang2025panoramic,
  title={Panoramic Video Generation with Pretrained Diffusion Models},
  author={Fang, Zixun and Zhu, Kai and Liu, Zhiheng and Liu, Yu and Zhai, Wei and Cao, Yang and Zha, Zheng-Jun},
  journal={arXiv preprint arXiv:2506.23513},
  year={2025}
}

@inproceedings{zhang2025panodit,
  title={PanoDit: Panoramic videos generation with diffusion transformer},
  author={Zhang, Muyang and Chen, Yuzhi and Xu, Rongtao and Wang, Changwei and Yang, JinMing and Meng, Weiliang and Guo, Jianwei and Zhao, Huihuang and Zhang, Xiaopeng},
  booktitle={Proceedings of the AAAI Conference on Artificial Intelligence},
  volume={39},
  number={10},
  pages={10040--10048},
  year={2025}
}

@article{huang2025dreamcube,
  title={DreamCube: 3D Panorama Generation via Multi-plane Synchronization},
  author={Huang, Yukun and Zhou, Yanning and Wang, Jianan and Huang, Kaiyi and Liu, Xihui},
  journal={arXiv preprint arXiv:2506.17206},
  year={2025}
}

@article{luo2025beyond,
  title={Beyond the Frame: Generating 360$^{\circ}$ Panoramic Videos from Perspective Videos},
  author={Luo, Rundong and Wallingford, Matthew and Farhadi, Ali and Snavely, Noah and Ma, Wei-Chiu},
  journal={arXiv preprint arXiv:2504.07940},
  year={2025}
}

@article{tan2024imagine360,
  title={Imagine360: Immersive 360 video generation from perspective anchor},
  author={Tan, Jing and Yang, Shuai and Wu, Tong and He, Jingwen and Guo, Yuwei and Liu, Ziwei and Lin, Dahua},
  journal={arXiv preprint arXiv:2412.03552},
  year={2024}
}

@inproceedings{ma2024vidpanos,
  title={VidPanos: Generative panoramic videos from casual panning videos},
  author={Ma, Jingwei and Lu, Erika and Paiss, Roni and Zada, Shiran and Holynski, Aleksander and Dekel, Tali and Curless, Brian and Rubinstein, Michael and Cole, Forrester},
  booktitle={SIGGRAPH Asia 2024 Conference Papers},
  pages={1--11},
  year={2024}
}

@inproceedings{kalischek2025cubediff,
  title={Cubediff: Repurposing diffusion-based image models for panorama generation},
  author={Kalischek, Nikolai and Oechsle, Michael and Manhardt, Fabian and Henzler, Philipp and Schindler, Konrad and Tombari, Federico},
  booktitle={The Thirteenth International Conference on Learning Representations},
  year={2025}
}

@article{lin2025one,
  title={One flight over the gap: A survey from perspective to panoramic vision},
  author={Lin, Xin and Ge, Xian and Zhang, Dizhe and Wan, Zhaoliang and Wang, Xianshun and Li, Xiangtai and Jiang, Wenjie and Du, Bo and Tao, Dacheng and Yang, Ming-Hsuan and others},
  journal={arXiv preprint arXiv:2509.04444},
  year={2025}
}

@article{he2024venhancer,
  title={Venhancer: Generative space-time enhancement for video generation},
  author={He, Jingwen and Xue, Tianfan and Liu, Dongyang and Lin, Xinqi and Gao, Peng and Lin, Dahua and Qiao, Yu and Ouyang, Wanli and Liu, Ziwei},
  journal={arXiv preprint arXiv:2407.07667},
  year={2024}
}

@inproceedings{huang2024vbench,
  title={Vbench: Comprehensive benchmark suite for video generative models},
  author={Huang, Ziqi and He, Yinan and Yu, Jiashuo and Zhang, Fan and Si, Chenyang and Jiang, Yuming and Zhang, Yuanhan and Wu, Tianxing and Jin, Qingyang and Chanpaisit, Nattapol and others},
  booktitle={Proceedings of the IEEE/CVF Conference on Computer Vision and Pattern Recognition},
  pages={21807--21818},
  year={2024}
}

@inproceedings{zhang2018unreasonable,
  title={The unreasonable effectiveness of deep features as a perceptual metric},
  author={Zhang, Richard and Isola, Phillip and Efros, Alexei A and Shechtman, Eli and Wang, Oliver},
  booktitle={Proceedings of the IEEE conference on computer vision and pattern recognition},
  pages={586--595},
  year={2018}
}

@inproceedings{radford2021learning,
  title={Learning transferable visual models from natural language supervision},
  author={Radford, Alec and Kim, Jong Wook and Hallacy, Chris and Ramesh, Aditya and Goh, Gabriel and Agarwal, Sandhini and Sastry, Girish and Askell, Amanda and Mishkin, Pamela and Clark, Jack and others},
  booktitle={International conference on machine learning},
  pages={8748--8763},
  year={2021},
  organization={PmLR}
}

@article{Qwen2.5-VL,
  title={Qwen2.5-VL Technical Report},
  author={Bai, Shuai and Chen, Keqin and Liu, Xuejing and Wang, Jialin and Ge, Wenbin and Song, Sibo and Dang, Kai and Wang, Peng and Wang, Shijie and Tang, Jun and Zhong, Humen and Zhu, Yuanzhi and Yang, Mingkun and Li, Zhaohai and Wan, Jianqiang and Wang, Pengfei and Ding, Wei and Fu, Zheren and Xu, Yiheng and Ye, Jiabo and Zhang, Xi and Xie, Tianbao and Cheng, Zesen and Zhang, Hang and Yang, Zhibo and Xu, Haiyang and Lin, Junyang},
  journal={arXiv preprint arXiv:2502.13923},
  year={2025}
}

@article{Qwen2-VL,
  title={Qwen2-VL: Enhancing Vision-Language Model's Perception of the World at Any Resolution},
  author={Wang, Peng and Bai, Shuai and Tan, Sinan and Wang, Shijie and Fan, Zhihao and Bai, Jinze and Chen, Keqin and Liu, Xuejing and Wang, Jialin and Ge, Wenbin and Fan, Yang and Dang, Kai and Du, Mengfei and Ren, Xuancheng and Men, Rui and Liu, Dayiheng and Zhou, Chang and Zhou, Jingren and Lin, Junyang},
  journal={arXiv preprint arXiv:2409.12191},
  year={2024}
}

@article{Qwen-VL,
  title={Qwen-VL: A Versatile Vision-Language Model for Understanding, Localization, Text Reading, and Beyond},
  author={Bai, Jinze and Bai, Shuai and Yang, Shusheng and Wang, Shijie and Tan, Sinan and Wang, Peng and Lin, Junyang and Zhou, Chang and Zhou, Jingren},
  journal={arXiv preprint arXiv:2308.12966},
  year={2023}
}

@inproceedings{esser2024scaling,
  title={Scaling rectified flow transformers for high-resolution image synthesis},
  author={Esser, Patrick and Kulal, Sumith and Blattmann, Andreas and Entezari, Rahim and M{\"u}ller, Jonas and Saini, Harry and Levi, Yam and Lorenz, Dominik and Sauer, Axel and Boesel, Frederic and others},
  booktitle={Forty-first international conference on machine learning},
  year={2024}
}

@article{lin2025autoregressive,
  title={Autoregressive Adversarial Post-Training for Real-Time Interactive Video Generation},
  author={Lin, Shanchuan and Yang, Ceyuan and He, Hao and Jiang, Jianwen and Ren, Yuxi and Xia, Xin and Zhao, Yang and Xiao, Xuefeng and Jiang, Lu},
  journal={arXiv preprint arXiv:2506.09350},
  year={2025}
}

@article{kim2024fifo,
  title={Fifo-diffusion: Generating infinite videos from text without training},
  author={Kim, Jihwan and Kang, Junoh and Choi, Jinyoung and Han, Bohyung},
  journal={Advances in Neural Information Processing Systems},
  volume={37},
  pages={89834--89868},
  year={2024}
}

@article{huang2025self,
  title={Self Forcing: Bridging the Train-Test Gap in Autoregressive Video Diffusion},
  author={Huang, Xun and Li, Zhengqi and He, Guande and Zhou, Mingyuan and Shechtman, Eli},
  journal={arXiv preprint arXiv:2506.08009},
  year={2025}
}

@inproceedings{akimoto2022diverse,
  title={Diverse plausible 360-degree image outpainting for efficient 3dcg background creation},
  author={Akimoto, Naofumi and Matsuo, Yuhi and Aoki, Yoshimitsu},
  booktitle={Proceedings of the IEEE/CVF Conference on Computer Vision and Pattern Recognition},
  pages={11441--11450},
  year={2022}
}

@article{chen2022text2light,
  title={Text2light: Zero-shot text-driven hdr panorama generation},
  author={Chen, Zhaoxi and Wang, Guangcong and Liu, Ziwei},
  journal={ACM Transactions on Graphics (TOG)},
  volume={41},
  number={6},
  pages={1--16},
  year={2022},
  publisher={ACM New York, NY, USA}
}

@inproceedings{cheng2022inout,
  title={Inout: Diverse image outpainting via gan inversion},
  author={Cheng, Yen-Chi and Lin, Chieh Hubert and Lee, Hsin-Ying and Ren, Jian and Tulyakov, Sergey and Yang, Ming-Hsuan},
  booktitle={Proceedings of the IEEE/CVF Conference on Computer Vision and Pattern Recognition},
  pages={11431--11440},
  year={2022}
}

@inproceedings{wang2022stylelight,
  title={Stylelight: Hdr panorama generation for lighting estimation and editing},
  author={Wang, Guangcong and Yang, Yinuo and Loy, Chen Change and Liu, Ziwei},
  booktitle={European conference on computer vision},
  pages={477--492},
  year={2022},
  organization={Springer}
}

@article{wu2022cross,
  title={Cross-view panorama image synthesis},
  author={Wu, Songsong and Tang, Hao and Jing, Xiao-Yuan and Zhao, Haifeng and Qian, Jianjun and Sebe, Nicu and Yan, Yan},
  journal={IEEE Transactions on Multimedia},
  volume={25},
  pages={3546--3559},
  year={2022},
  publisher={IEEE}
}

@article{xia2025panowan,
  title={PanoWan: Lifting Diffusion Video Generation Models to 360 $\{$$\backslash$deg$\}$ with Latitude/Longitude-aware Mechanisms},
  author={Xia, Yifei and Weng, Shuchen and Yang, Siqi and Liu, Jingqi and Zhu, Chengxuan and Teng, Minggui and Jia, Zijian and Jiang, Han and Shi, Boxin},
  journal={arXiv preprint arXiv:2505.22016},
  year={2025}
}

@article{karras2022elucidating,
  title={Elucidating the design space of diffusion-based generative models},
  author={Karras, Tero and Aittala, Miika and Aila, Timo and Laine, Samuli},
  journal={Advances in neural information processing systems},
  volume={35},
  pages={26565--26577},
  year={2022}
}

@article{ho2022classifier,
  title={Classifier-free diffusion guidance},
  author={Ho, Jonathan and Salimans, Tim},
  journal={arXiv preprint arXiv:2207.12598},
  year={2022}
}

@inproceedings{rombach2022high,
  title={High-resolution image synthesis with latent diffusion models},
  author={Rombach, Robin and Blattmann, Andreas and Lorenz, Dominik and Esser, Patrick and Ommer, Bj{\"o}rn},
  booktitle={Proceedings of the IEEE/CVF conference on computer vision and pattern recognition},
  pages={10684--10695},
  year={2022}
}

@article{blattmann2023stable,
  title={Stable video diffusion: Scaling latent video diffusion models to large datasets},
  author={Blattmann, Andreas and Dockhorn, Tim and Kulal, Sumith and Mendelevitch, Daniel and Kilian, Maciej and Lorenz, Dominik and Levi, Yam and English, Zion and Voleti, Vikram and Letts, Adam and others},
  journal={arXiv preprint arXiv:2311.15127},
  year={2023}
}

@article{ho2020denoising,
  title={Denoising diffusion probabilistic models},
  author={Ho, Jonathan and Jain, Ajay and Abbeel, Pieter},
  journal={Advances in neural information processing systems},
  volume={33},
  pages={6840--6851},
  year={2020}
}

@inproceedings{blattmann2023align,
  title={Align your latents: High-resolution video synthesis with latent diffusion models},
  author={Blattmann, Andreas and Rombach, Robin and Ling, Huan and Dockhorn, Tim and Kim, Seung Wook and Fidler, Sanja and Kreis, Karsten},
  booktitle={Proceedings of the IEEE/CVF Conference on Computer Vision and Pattern Recognition},
  pages={22563--22575},
  year={2023}
}

@article{teng2025magi,
  title={MAGI-1: Autoregressive Video Generation at Scale},
  author={Teng, Hansi and Jia, Hongyu and Sun, Lei and Li, Lingzhi and Li, Maolin and Tang, Mingqiu and Han, Shuai and Zhang, Tianning and Zhang, WQ and Luo, Weifeng and others},
  journal={arXiv preprint arXiv:2505.13211},
  year={2025}
}

@inproceedings{yin2025slow,
  title={From slow bidirectional to fast autoregressive video diffusion models},
  author={Yin, Tianwei and Zhang, Qiang and Zhang, Richard and Freeman, William T and Durand, Fredo and Shechtman, Eli and Huang, Xun},
  booktitle={Proceedings of the Computer Vision and Pattern Recognition Conference},
  pages={22963--22974},
  year={2025}
}

@article{zhang2025packing,
  title={Packing input frame context in next-frame prediction models for video generation},
  author={Zhang, Lvmin and Agrawala, Maneesh},
  journal={arXiv preprint arXiv:2504.12626},
  year={2025}
}

@article{chen2024diffusion,
  title={Diffusion forcing: Next-token prediction meets full-sequence diffusion},
  author={Chen, Boyuan and Mart{\'\i} Mons{\'o}, Diego and Du, Yilun and Simchowitz, Max and Tedrake, Russ and Sitzmann, Vincent},
  journal={Advances in Neural Information Processing Systems},
  volume={37},
  pages={24081--24125},
  year={2024}
}

@article{guo2025long,
  title={Long context tuning for video generation},
  author={Guo, Yuwei and Yang, Ceyuan and Yang, Ziyan and Ma, Zhibei and Lin, Zhijie and Yang, Zhenheng and Lin, Dahua and Jiang, Lu},
  journal={arXiv preprint arXiv:2503.10589},
  year={2025}
}

@article{peebles2023scalable,
  title={Scalable Diffusion Models with Transformers},
  author={Peebles, William and Xie, Saining},
  journal={arXiv preprint arXiv:2303.12345},
  year={2023}
}

@article{liu2025rolling,
  title={Rolling Forcing: Autoregressive Long Video Diffusion in Real Time},
  author={Liu, Kunhao and Hu, Wenbo and Xu, Jiale and Shan, Ying and Lu, Shijian},
  journal={arXiv preprint arXiv:2509.25161},
  year={2025}
}

@inproceedings{cao2023ntire,
  title={Ntire 2023 challenge on 360deg omnidirectional image and video super-resolution: Datasets, methods and results},
  author={Cao, Mingdeng and Mou, Chong and Yu, Fanghua and Wang, Xintao and Zheng, Yinqiang and Zhang, Jian and Dong, Chao and Li, Gen and Shan, Ying and Timofte, Radu and others},
  booktitle={Proceedings of the IEEE/CVF conference on computer vision and pattern recognition},
  pages={1731--1745},
  year={2023}
}

@article{heusel2017gans,
  title={Gans trained by a two time-scale update rule converge to a local nash equilibrium},
  author={Heusel, Martin and Ramsauer, Hubert and Unterthiner, Thomas and Nessler, Bernhard and Hochreiter, Sepp},
  journal={Advances in neural information processing systems},
  volume={30},
  year={2017}
}

@article{unterthiner2018towards,
  title={Towards accurate generative models of video: A new metric \& challenges},
  author={Unterthiner, Thomas and Van Steenkiste, Sjoerd and Kurach, Karol and Marinier, Raphael and Michalski, Marcin and Gelly, Sylvain},
  journal={arXiv preprint arXiv:1812.01717},
  year={2018}
}

@article{zhuang2025flashvsr,
  title={FlashVSR: Towards Real-Time Diffusion-Based Streaming Video Super-Resolution},
  author={Zhuang, Junhao and Guo, Shi and Cai, Xin and Li, Xiaohui and Liu, Yihao and Yuan, Chun and Xue, Tianfan},
  journal={arXiv preprint arXiv:2510.12747},
  year={2025}
}

@inproceedings{lu2024autoregressive,
  author       = {Zhuqiang Lu and
                  Kun Hu and
                  Chaoyue Wang and
                  Lei Bai and
                  Zhiyong Wang},
  editor       = {Michael J. Wooldridge and
                  Jennifer G. Dy and
                  Sriraam Natarajan},
  title        = {Autoregressive Omni-Aware Outpainting for Open-Vocabulary 360-Degree
                  Image Generation},
  booktitle    = {Thirty-Eighth {AAAI} Conference on Artificial Intelligence, {AAAI}
                  2024, February 20-27, 2024, Vancouver,
                  Canada},
  pages        = {14211--14219},
  publisher    = {{AAAI} Press},
  year         = {2024},
  url          = {https://doi.org/10.1609/aaai.v38i13.29332},
  doi          = {10.1609/AAAI.V38I13.29332},
  bibsource    = {dblp computer science bibliography, https://dblp.org}
}
}
\end{document}